\documentclass{article}

\PassOptionsToPackage{numbers, compress}{natbib}

\usepackage[final]{bdl_2018}

\usepackage[utf8]{inputenc} 
\usepackage[T1]{fontenc}    
\usepackage{hyperref}       
\usepackage{url}            
\usepackage{booktabs}       
\usepackage{amsfonts}       
\usepackage{nicefrac}       
\usepackage{microtype}      
\usepackage{algorithm,algpseudocode}
\usepackage{amsmath}
\usepackage{graphicx}
\usepackage{subcaption}
\graphicspath{./}

\title{Sampling-based Bayesian Inference with \\gradient uncertainty}

\author{
  Chanwoo Park \quad Jae Myung Kim \quad Seok Hyeon Ha \quad Jungwoo Lee\\
  Department of Electrical and Computer Engineering\\
  Seoul National University\\
  \texttt{\{cpark, goldkim92, hash1108\}@cml.snu.ac.kr, junglee@snu.ac.kr} \\
}

\begin{document}

\maketitle

\begin{abstract}
  Deep neural networks(NNs) have achieved impressive performance, often exceed human performance on many computer vision tasks. However, one of the most challenging issues that still remains is that NNs are overconfident in their predictions, which can be very harmful when this arises in safety critical applications. In this paper, we show that predictive uncertainty can be efficiently estimated when we incorporate the concept of gradients uncertainty into posterior sampling. The proposed method is tested on two different datasets, MNIST for in-distribution confusing examples and notMNIST for out-of-distribution data. We show that our method is able to efficiently represent predictive uncertainty on both datasets.

\end{abstract}

\section{Introduction}

Recent deep neural networks (NN) methods have achieved human-level or superhuman performance at various tasks in computer vision, natural language processing, and robotics.
But they often fail to estimate predictive uncertainty which can be crucial for some tasks such as medical diagnosis or autonomous driving, and tend to be overconfident even when their predictions are incorrect. 
In order for these critical applications to be successfully employed, they must be able to provide how certain they are about cancer detection and road sign recognition  \cite{amodei2016concrete}.
 Bayesian NNs, which find a posterior distribution over network parameters, are the state-of-the-art methods for estimating predictive uncertainty \cite{neal2012bayesian, mackay1992practical}.

\subsection{Bayes's Theorem}
\begin{equation}
p(\theta |\small D) = \frac{p(D |\small \theta) \, p(\theta)}{p(D)} = \frac{p(D|\small \theta) \, p(\theta)}{\int{}p(D |\small \theta) \, p(\theta)\,d\theta}    
\end{equation}

$p(D|\small\theta)$ is the likelihood of the occurrence of dataset $D$ given a model with parameters $\theta$, $p(\theta)$ the prior, and $p(D)$ the data distribution.
Bayesian NNs compute the posterior distribution over the parameters to estimate predictive uncertainty. 
However exact Bayesian inference is not feasible for NN since computing marginal likelihoods in high dimension is analytically intractable, thus we need to approximate the inference problem.

\subsection{Related Works}
There have been two different major approaches for the approximation.
One is variational Bayesian methods, which approximate Bayesian inference by introducing simpler, tractable distribution $q(\theta)$ to approximate posterior distribution \cite{hinton1993keeping, barber1998ensemble, graves2011practical,blundell2015weight, gal2016dropout}. 
This method minimizes the Kullback-Leibler (KL) divergence between $p$ and $q$, $KL(q(\theta)|| p(\theta|D))$, which is optimized by maximizing the evidence lower bound (ELBO). 
On the other hand, Markov Chain Monte Carlo (MCMC) methods have been successfully applied to Bayesian NNs.
MCMC is non-parametric and asymptotically exact which iteratively draw samples from unknown true distribution to approximate expectation \cite{chib1995understanding, dellaportas2002bayesian, hanson2001markov}. 
However, traditional MCMC methods require the full dataset in each iteration to generate proposals and calculate the acceptance probability which make them prohibitively expensive for large datasets.
\par
Stochastic gradient MCMC (SG-MCMC) has gained keen interest recently which uses minibatches of the data to generate samples and ignore the acceptance step, thus scales well to large datasets. 
Welling \& Teh(2011) developed stochastic-gradient Langevin dynamics(SGLD) by incorporating Langevin dynamics into stochastic optimization to insert adaptively scaled Gaussian noise, which is the first sampling algorithm based on stochastic gradients \cite{welling2011bayesian}. 
Chen et al.(2014) suggested Stochastic gradient Hamiltonian Monte Carlo (SGHMC) where they introduced auxiliary momentum term to rapidly explore the parameter space \cite{chen2014stochastic}.
Recent studies have been proposed to improve convergence and sampling efficiency of SGLD and SGHMC for the past years \cite{ahn2012bayesian, li2016preconditioned, patterson2013stochastic, chen2016bridging}.

\section{Our methods}

In this paper we try to sample from an approximate posterior distribution using gradient uncertainty.
Let $X^{(l)} = (X_{l,1},X_{l,2}, ...,X_{l,m})$ denotes the sequence of gradient vector of $i_{th}$ example in $l_{th}$ minibatch where $X_{l,i} =  \nabla_{\theta} J(\theta; x^{(i)}, y^{(i)})$ and $m$ is the minibatch size.
We define a new measure to quantify gradient uncertainty. 

\begin{equation}\label{equ_gradientuncertainty}
Gradient\: Uncertainty = \sum_{i\neq j}\left< X_i, X_j \right>
\end{equation}

To the extent of our knowledge, this is the first work using stochastic gradient uncertainty for Bayesian sampling algorithm. 
The main idea is that we use gradient uncertainty as an indicator that the parameters are near the local optimum.
The proposed sampling method is outlined in Alg \ref{alg:alg1}. 

\begin{algorithm}
\caption{Sampling algorithm}
\label{alg:alg1}
\begin{algorithmic}[1]
\State \textbf{Initialize: } Random $\theta_1$
\For{$l$ = 1,2,...,T}

    \For{$i$ = 1,2,...,m} \Comment{m : minibatch size}
    \State $X_{l,i} \gets$ $\nabla_{\theta} J(\theta; x^{(i)}, y^{(i)})$
    \EndFor
    \State $\Tilde{X_l} \gets$ $\frac{1}{m} \sum_{i=1}^{m} X_{l,i}$
    \State $gradient\: uncertainty \gets$ $\sum_{i\neq j}\left< X_i, X_j \right>$
            \If{$gradient\: uncertainty < threshold$}
                \State Sample $\theta_{l}$ 
                \State $\theta_{l+1} \gets$ $\theta_{l}-\epsilon\Tilde{X_l}$ + $N(0,\sigma^{2} )$ \Comment{$\epsilon$ : learning rate}
            \Else
                \State $\theta_{l+1} \gets$ $\theta_{l}-\epsilon\Tilde{X_l}$
            \EndIf
    \EndFor
\end{algorithmic}
\end{algorithm}

When the algorithm reaches local extrema, the exact gradient should be either zero or very close to zero.
However, because of unbiased but noisy stochastic gradient, we assume the stochastic gradient near local modes follows zero-mean Gaussian distribution. 
Since stochastic gradients from a single training example, $X_{i}$, are randomly scattered around zero in each dimension of the gradient vector, the sum of inner product between different gradient vectors in a minibatch, Eq. (\ref{equ_gradientuncertainty}), gives a sufficiently small value, whereas the sum of inner product can be large if the gradient vectors are not following zero-mean Gaussian but pointing in the similar direction. \par
We evaluate \emph{gradient uncertainty} from the minibatch and sample parameters when it is lower than a certain threshold, meaning that we assume it has reached near local mode. 
That is, the magnitude of mean of gradients is sufficiently small and directions of gradients are diverse.
It is computationally efficient because we calculate \emph{gradient uncertainty} only from data subsets. \par
In most NN, the loss functions are non-convex and thus gradient-based optimization might get stuck in a local mode. 
We add random Gaussian noise to a proposal every time we sample a set of parameters, so that our method can escape from local mode and explore the parameter space.

\section{Experiments}
We conduct two sets of experiments with MNIST \cite{mnistlecun2010mnist} and notMNIST datasets for uncertainty estimation. The notMNIST dataset contains front glyphs for the 10 class letters(A-J), which can be treated as out-of-distribution data for the networks trained on original MNIST.
 The notMNIST dataset consists of 19K hand-cleaned instances and 500K uncleaned instances \cite{notmnistbulatov2011notmnist}. \par
 
 \subsection{MNIST : Confusing examples}
\begin{figure}[t]
 \centering
\begin{subfigure}{0.20\textwidth}
\includegraphics[width=0.9\linewidth]{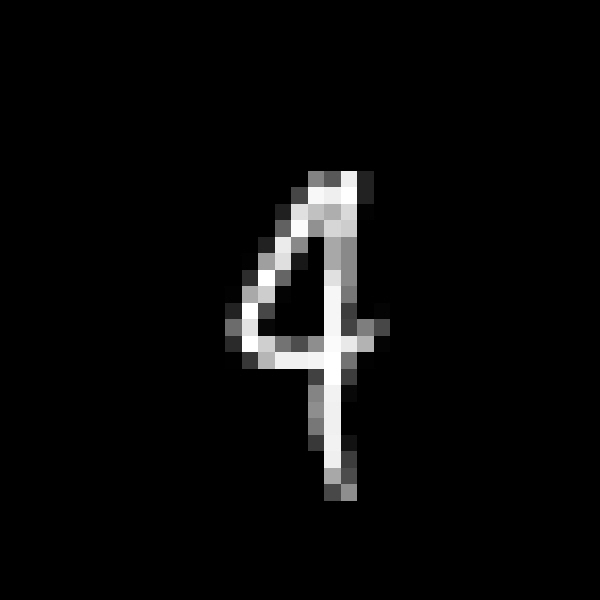}
\caption*{Digit 4}
\end{subfigure}
\begin{subfigure}{0.26\textwidth}
\includegraphics[width=0.9\linewidth]{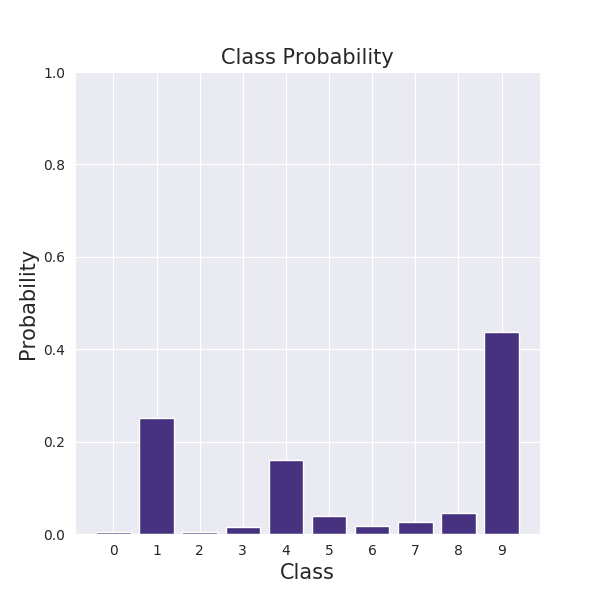}
\caption*{}
\end{subfigure}
 \begin{subfigure}{0.26\textwidth}
\includegraphics[width=0.9\linewidth]{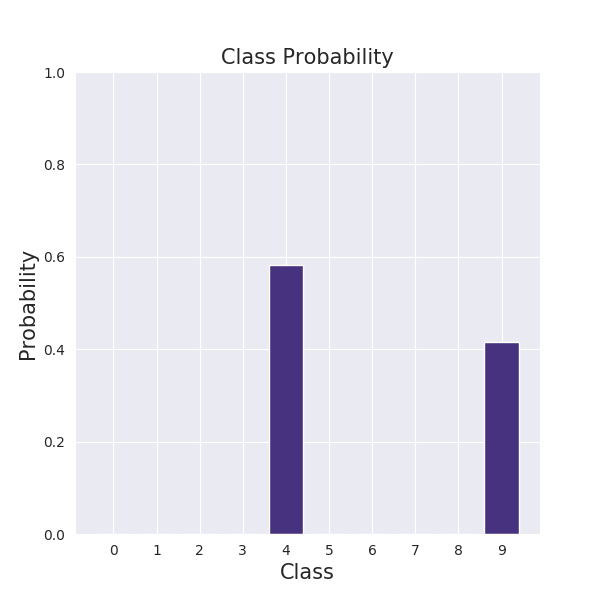}
\caption*{}
\end{subfigure}
\begin{subfigure}{0.26\textwidth}
\includegraphics[width=0.9\linewidth]{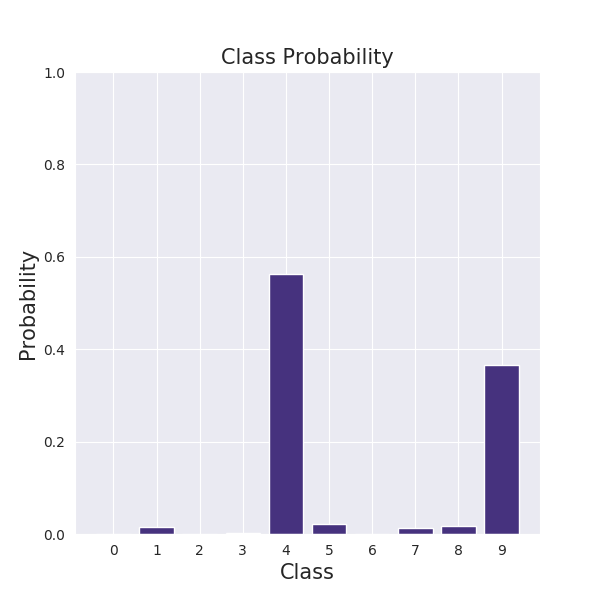}
\caption*{}
\end{subfigure}

\begin{subfigure}{0.20\textwidth}
\includegraphics[width=0.9\linewidth]{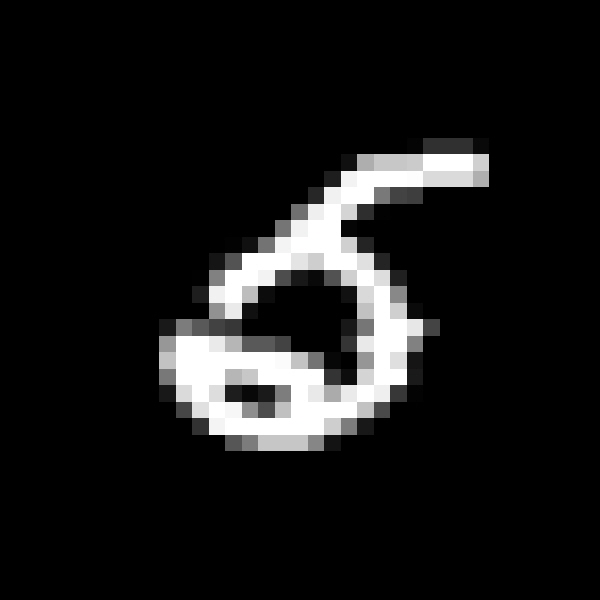}
\caption*{Digit 5}
\end{subfigure}
\begin{subfigure}{0.26\textwidth}
\includegraphics[width=0.9\linewidth]{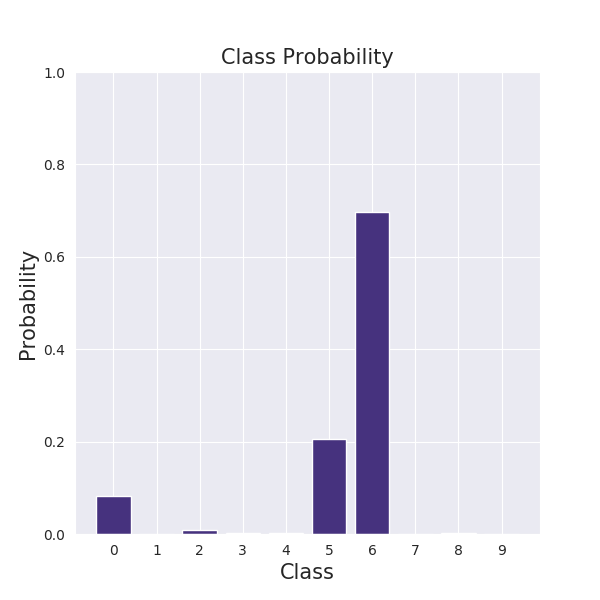}
\caption*{}
\end{subfigure}
 \begin{subfigure}{0.26\textwidth}
\includegraphics[width=0.9\linewidth]{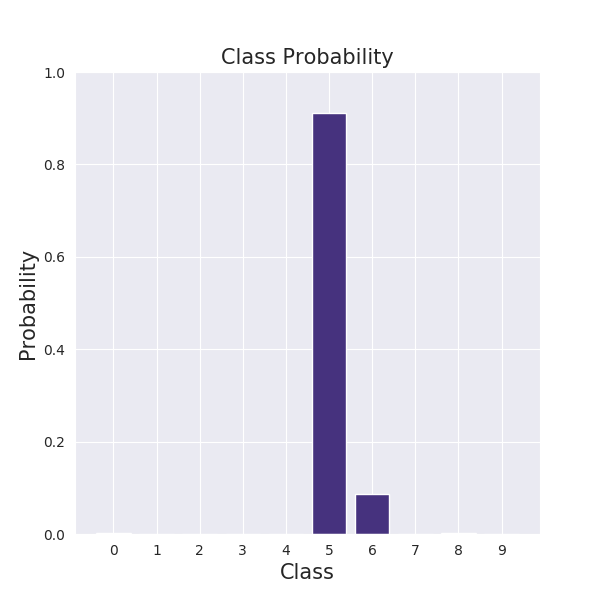}
\caption*{}
\end{subfigure}
\begin{subfigure}{0.26\textwidth}
\includegraphics[width=0.9\linewidth]{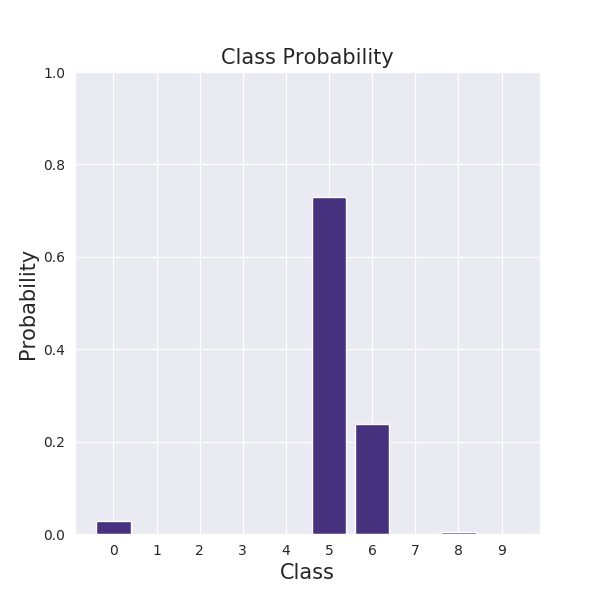}
\caption*{}
\end{subfigure}

\begin{subfigure}{0.20\textwidth}
\includegraphics[width=0.9\linewidth]{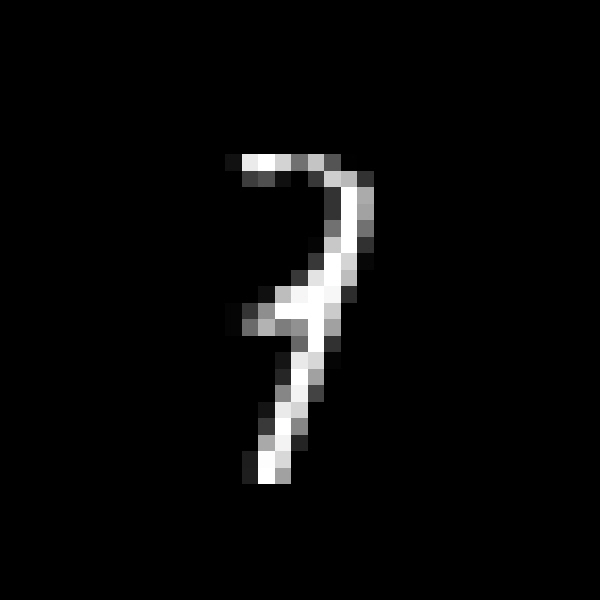}
\caption*{Digit 7}
\end{subfigure}
\begin{subfigure}{0.26\textwidth}
\includegraphics[width=0.9\linewidth]{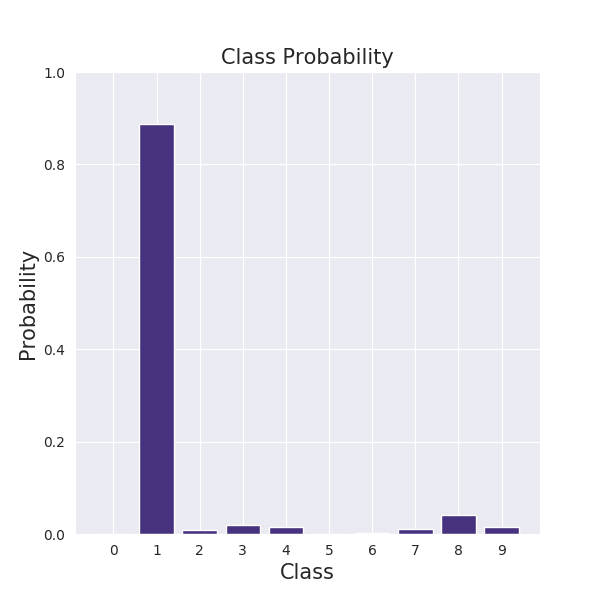}
\caption{SGLD}
\end{subfigure}
 \begin{subfigure}{0.26\textwidth}
\includegraphics[width=0.9\linewidth]{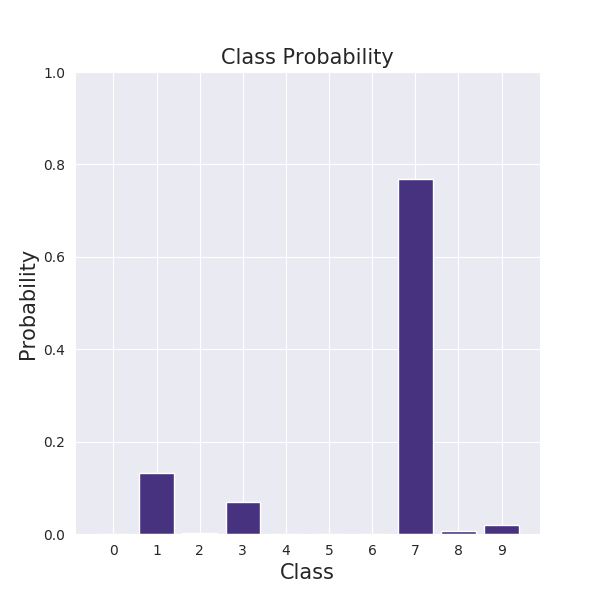}
\caption{Dropout}
\end{subfigure}
\begin{subfigure}{0.26\textwidth}
\includegraphics[width=0.9\linewidth]{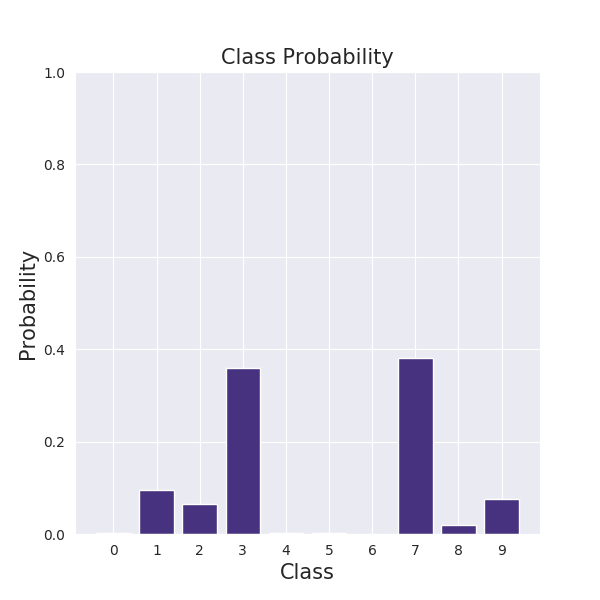}
\caption{Ours}
\end{subfigure}

\caption{MNIST Confusing examples}
\label{fig:image2}
\end{figure}

In the first experiment, we train networks on MNIST and compare the ability of three different methods to quantify uncertainty when confusing examples are given. We take the same five-layer convolutional NN which are trained on MNIST, and then evaluate the predictive uncertainty with 60 Monte Carlo samples. The results are compared with Dropout Bayesian Approximation \cite{gal2016dropout} and SGLD \cite{welling2011bayesian}. 
This is shown in Figure \ref{fig:image2}.

In our approach the predictive results of given confusing digits are diverse and the probability is concentrated on the most likely candidates, i.e. in our cases '4', '5', '7' can be misinterpreted as '9', '6', '3' respectively. On the other hand, other methods show an overconfident prediction by assigning higher probability to incorrect classes.\par

\subsection{notMNIST : Out-of-distribution data}

\begin{figure}[t]
 \centering
\begin{subfigure}{0.20\textwidth}
\includegraphics[width=0.9\linewidth]{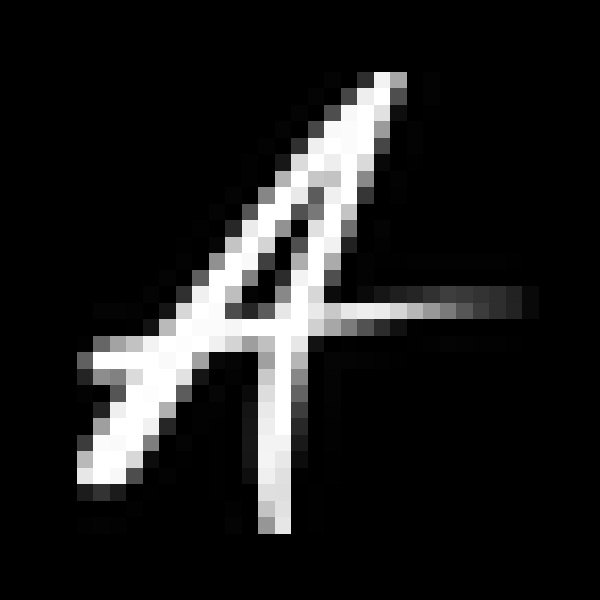}
\caption*{Letter A}
\end{subfigure}
\begin{subfigure}{0.26\textwidth}
\includegraphics[width=0.9\linewidth]{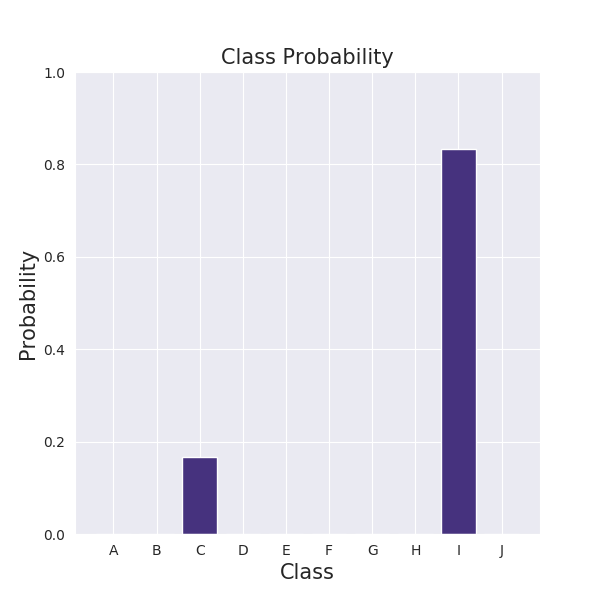}
\caption*{}
\end{subfigure}
 \begin{subfigure}{0.26\textwidth}
\includegraphics[width=0.9\linewidth]{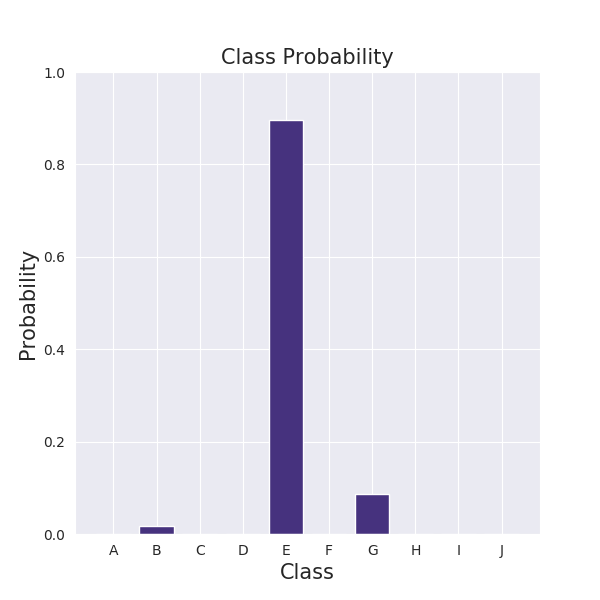}
\caption*{}
\end{subfigure}
\begin{subfigure}{0.26\textwidth}
\includegraphics[width=0.9\linewidth]{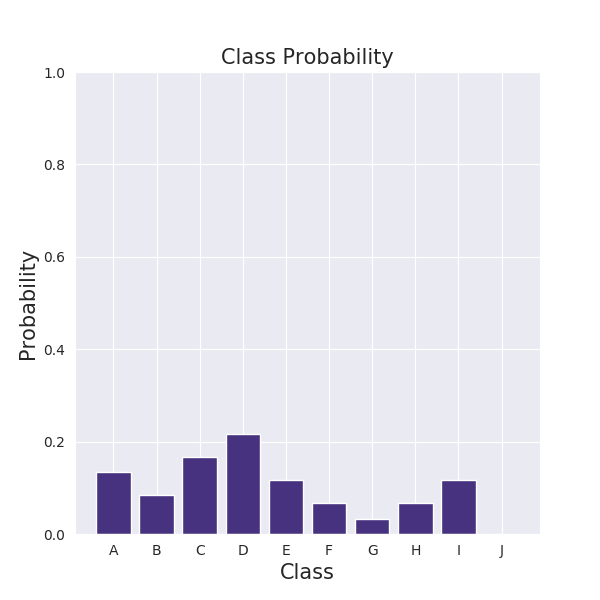}
\caption*{}
\end{subfigure}

\begin{subfigure}{0.20\textwidth}
\includegraphics[width=0.9\linewidth]{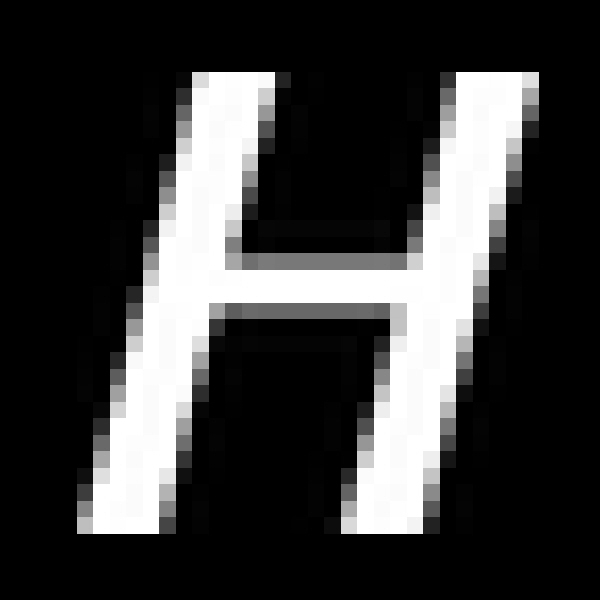}
\caption*{Letter H}
\end{subfigure}
\begin{subfigure}{0.26\textwidth}
\includegraphics[width=0.9\linewidth]{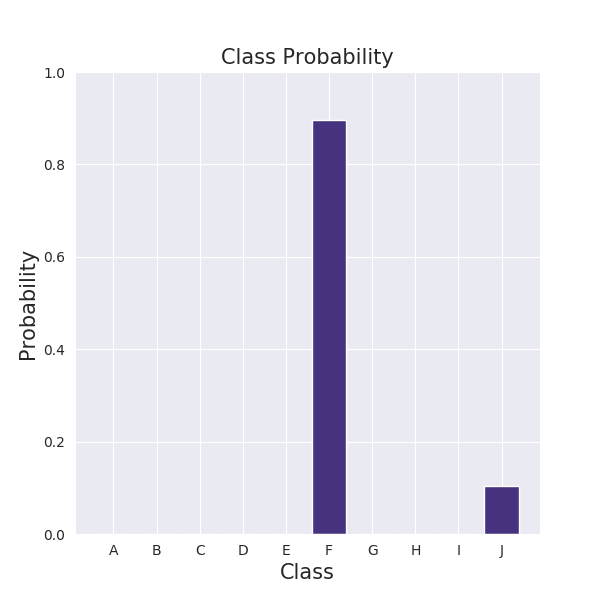}
\caption{SGLD}
\end{subfigure}
 \begin{subfigure}{0.26\textwidth}
\includegraphics[width=0.9\linewidth]{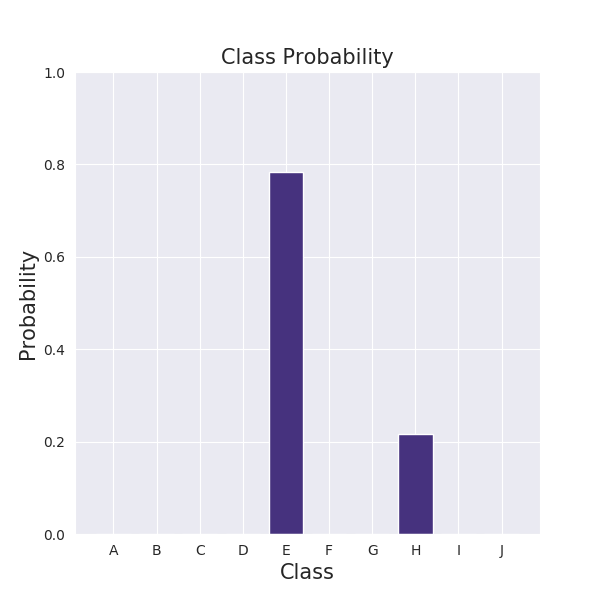}
\caption{Dropout}
\end{subfigure}
\begin{subfigure}{0.26\textwidth}
\includegraphics[width=0.9\linewidth]{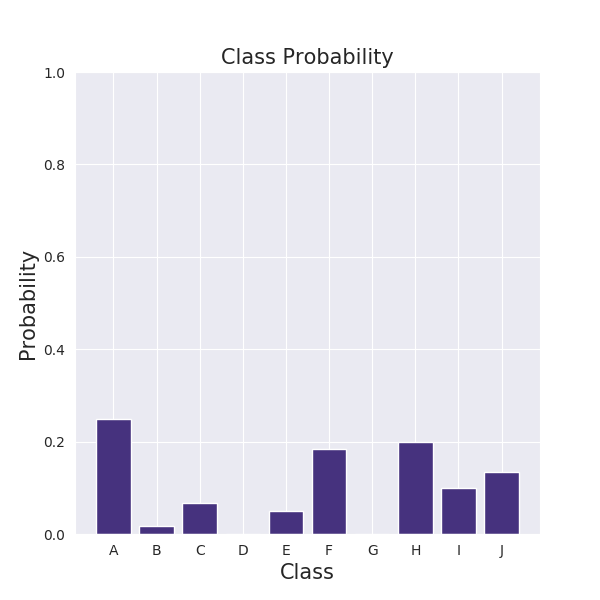}
\caption{Ours}
\end{subfigure}

\caption{notMNIST Out-of-distribution examples}
\label{fig:image3}
\end{figure}

We further experimented with the same networks trained on MNIST to evaluate the uncertainty estimation on out-of-distribution data, which is notMNIST dataset in our case. 
Figure \ref{fig:image3} shows the comparison results with the same baseline methods.
It is shown that the proposed approach generates a fairly flat predictive posterior, while others are rather sharp.
This result shows that our method captures reliable uncertainty when classifying unseen data.\par

To compare the effectiveness of uncertainty estimation on out-of-distribution data, we use the entropy of the predictive distribution, $H(Y|X)$.
The predictive distribution for a given image X and the entropy can be expressed as: 
\begin{equation}
p(Y|X) = \int{}p(Y|X,\theta)p(\theta|D)d\theta
\end{equation}
\begin{equation}
H(Y|X) = -\int{}p(y|X) \ln(p(y|X))dy
\end{equation}
We approximate $H(Y|X)$ using 60 Monte Carlo samples over the notMNIST dataset.
Entropy comparison with baseline methods can be seen in Table \ref{table:1}.
Our method shows higher predictive entropy on out-of-distribution data, i.e., our model gives less confident predictions on them. 
In both sets of experiments, we observe that our approach well represents predictive uncertainty while SGLD and Dropout generally produce overconfident probabilities.

\begin{table}[h]
\centering
\begin{tabular}{|c|c|c|c|}
\hline
                                                                                 & \textbf{Ours}  & \textbf{SGLD} & \textbf{Dropout} \\ \hline
\begin{tabular}[c]{@{}c@{}}In-distribution \\ Test Accuracy (\%)\end{tabular}    & 98.05          & 95.34         & 99.39            \\ \hline
\begin{tabular}[c]{@{}c@{}}Out-of-distribution\\ Predictive Entropy\end{tabular} & \textbf{1.355} & 0.107         & 0.665            \\ \hline
\end{tabular}
\caption{Entropy Comparison}
\label{table:1}
\end{table}



\section{Conclusion}
Overconfident prediction is a major obstacle for many deep learning architectures to be deployed in safety critical applications. This paper presents a new methodology for uncertainty estimation which uses stochastic gradient uncertainty as an indicator to sample. We experimented our model on MNIST and notMNIST datasets. In both cases, we have shown that our method efficiently sample from posterior using stochastic gradients based computation. It is an important future work to compare quality of uncertainty estimates with other state-of-the-art SG-MCMC and variational methods. Also, applying adaptive learning rate based on gradient uncertainty when performing SG-MCMC will be another line of future work. 

\subsubsection*{Acknowledgments}
This work is in part supported by Basic Science Research Program (NRF-2017R1A2B2007102) through NRF funded by MSIP, Technology Innovation Program (10051928) funded by MOTIE, Bio-Mimetic Robot Research Center funded by DAPA (UD130070ID), INMAC, and BK21-plus.

\bibliographystyle{plain}
\bibliography{cp_paper.bib}

\end{document}